\documentclass[journal,twoside,web]{ieeecolor}
\usepackage{tmi}
\usepackage{cite}
\usepackage{amsmath,amssymb,amsfonts}
\usepackage{algorithmic}
\usepackage{graphicx}
\usepackage{textcomp}
\usepackage{hyperref}
\usepackage{amssymb}
\usepackage[table]{xcolor} 
\usepackage{booktabs} 
\usepackage{multirow} 
\usepackage{rotating} 
\usepackage{amsmath}
\usepackage{soul}
\usepackage{subfig}
\definecolor{oran_tab}{RGB}{252, 242, 237}
\definecolor{blue_tab}{RGB}{227, 240, 251}
\definecolor{green_tab}{RGB}{240, 255, 240}
\def\BibTeX{{\rm B\kern-.05em{\sc i\kern-.025em b}\kern-.08em
    T\kern-.1667em\lower.7ex\hbox{E}\kern-.125emX}}
\begin{document}
\title{NeuroMamba: Multi-Perspective Feature Interaction with Visual Mamba 
	for Neuron Segmentation}

\author{Liuyun Jiang, Yizhuo Lu, Yanchao Zhang, Jiazheng Liu, and Hua Han
	\thanks{This work was supported by the grants from the STI 2030-Major 
	Projects (2021ZD0204500, 2021ZD0204503), the National Natural Science 
	Foundation of China (32171461). }
	\thanks{Liuyun Jiang, Yizhuo Lu, Yanchao Zhang and Hua Han are
		with State Key Laboratory of Brain
		Cognition and Brain-Inspired Intelligence Technology,
		Institute of Automation, Chinese Academy of Sciences, Beijing, China, 
		and also with School of Future Technology, University of Chinese 
		Academy of Sciences, Beijing, China (e-mail: jiangliuyun2023@ia.ac.cn; 
		luyizhuo2023@ia.ac.cn; zhangyanchao2021@ia.ac.cn; hua.han@ia.ac.cn).}
	\thanks{Jiazheng Liu is with State Key Laboratory of Brain Cognition and 
	Brain-inspired Intelligence Technology, Institute of Automation, Chinese 
	Academy of Sciences, Beijing, China (e-mail: liujiazheng2018@ia.ac.cn).}
}

\maketitle

\begin{abstract}
Neuron segmentation is the cornerstone of reconstructing comprehensive neuronal 
connectomes, which is essential for deciphering the functional organization of 
the brain. The irregular morphology and densely intertwined structures of 
neurons make this task particularly challenging. Prevailing CNN-based methods 
often fail to resolve ambiguous boundaries due to the lack of long-range 
context, whereas Transformer-based methods suffer from boundary imprecision 
caused by the loss of voxel-level details during patch partitioning. To address 
these limitations, we propose NeuroMamba, a multi-perspective framework that 
exploits the linear complexity of Mamba to enable patch-free global modeling 
and synergizes this with complementary local feature modeling, thereby 
efficiently capturing long-range dependencies while meticulously preserving 
fine-grained voxel details. Specifically, we design a channel-gated Boundary 
Discriminative Feature Extractor (BDFE) to enhance local morphological cues. 
Complementing this, we introduce the Spatial Continuous Feature Extractor 
(SCFE), which integrates a resolution-aware scanning mechanism into the Visual 
Mamba architecture to adaptively model global dependencies across varying data 
resolutions. Finally, a cross-modulation mechanism synergistically fuses these 
multi-perspective features. Our method demonstrates state-of-the-art 
performance across four public EM datasets, validating its exceptional 
adaptability to both anisotropic and isotropic resolutions. The source code 
will be made publicly available.

\end{abstract}


\section{Introduction}
\IEEEPARstart{N}{euron} segmentation serves as the cornerstone of connectomics, 
enabling the reconstruction of synapse-level neural circuits to decipher the 
functional organization of the brain \cite{a:5} and advance our understanding 
of biological neural functions \cite{a:6,a:7,a:8}. Currently, volumetric 
electron microscopy (EM) at nanometer resolution remains the premier technique 
capable of capturing the intricate morphology of neurons and individual 
synapses within brain tissue \cite{a:9,a:10}. However, even small volumes of 
neural tissue produce vast amounts of data at this resolution 
\cite{a:11,a:12,a:13}, rendering manual reconstruction prohibitively 
time-consuming and labor-intensive, creating an urgent demand for efficient 
automated segmentation methods.

\begin{figure}[t]
	\centering
	\includegraphics[width=1.0\columnwidth]{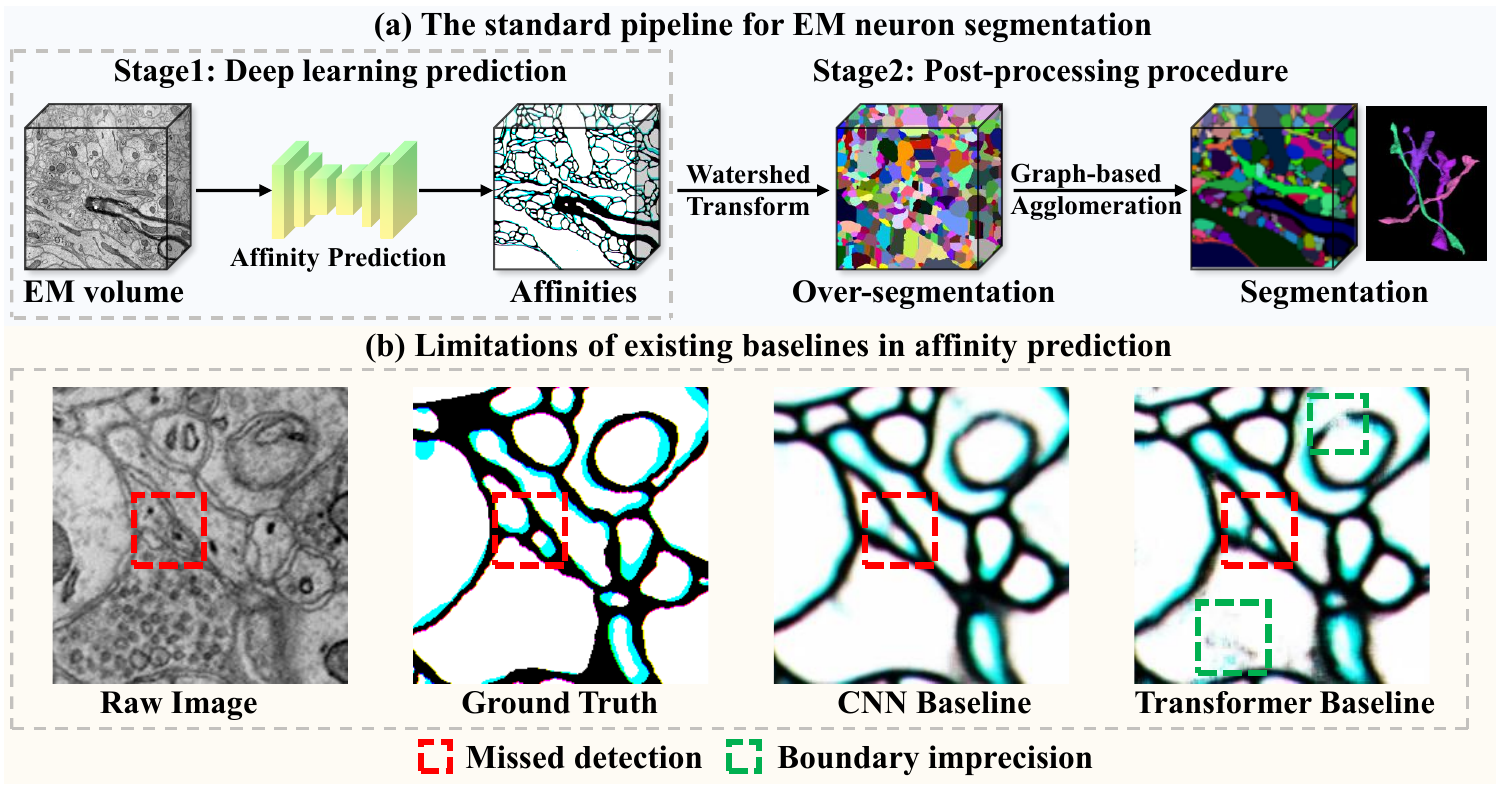} 
	\caption{Illustration of the standard EM neuron segmentation pipeline and 
	limitations of existing baselines. 
		(a) The standard two-stage pipeline: deep learning-based affinity 
		prediction followed by post-processing. 
		(b) Complementary failures in affinity prediction. The CNN baseline 
		misses ambiguous membranes due to limited context (red box), while the 
		Transformer baseline suffers from boundary imprecision due to the loss 
		of voxel details caused by patch partitioning (green boxes).}
	\label{fig1}
\end{figure}

To overcome these obstacles, automated methods based on deep learning have been 
extensively developed~\cite{superhuman,mala,ffn,3dmetric}. Currently, the 
prevailing paradigm follows a two-stage boundary detection framework, as 
illustrated in Fig.~\ref{fig1}(a). In this pipeline, a deep neural network 
first predicts voxel-wise affinities~\cite{aff}, which are subsequently 
processed by 
post-processing algorithms~\cite{multicut,mala,deepmulticut} to generate the 
final instance segmentation. Despite 
these advancements, achieving accurate reconstruction remains a significant 
hurdle. The complex morphology, dense branching, and diverse textures of 
neurons demand a model that can capture both fine-grained local details for 
precise boundary detection and long-range spatial dependencies for structural 
continuity. This dual requirement exposes the fundamental limitations of 
prevailing deep learning architectures, as highlighted in Fig.~\ref{fig1}(b). 
On the one hand, Convolutional Neural Networks (CNNs)~\cite{superhuman,mala}, 
while effective at extracting local features, are constrained by their inherent 
locality. Their limited receptive fields often fail to capture sufficient 
context to resolve ambiguous boundaries (as seen in the red box), making them 
prone to topological errors such as erroneous splits. On the other hand, 
Transformer-based methods~\cite{self,appearance,UNETR,SwinUNETR} leverage 
global context via the attention mechanism to better resolve ambiguous regions. 
However, this patch-based paradigm often neglects crucial correlations among 
voxels within each patch, leading to boundary imprecision (as seen in the green 
box). Furthermore, standard Transformers often struggle to adapt to the 
anisotropic resolution of 3D EM data, where significant discrepancies between 
transverse and axial resolutions complicate voxel affinity modeling. 
Consequently, a clear and urgent need exists for a novel architecture capable 
of efficiently and holistically modeling both global continuity and local 
details without the inherent drawbacks of existing approaches.

The Mamba architecture~\cite{mamba}, built upon state space models 
(SSMs)~\cite{ssm}, has recently emerged as a compelling successor to 
Transformers for efficient long-sequence modeling. Mamba distinguishes itself 
by its ability to capture long-range dependencies with linear complexity. This 
unique attribute offers a paradigm-shifting advantage for neuron segmentation: 
it enables the modeling of global voxel dependencies across entire volumetric 
blocks without the need for patch partitioning. By processing the raw volume 
directly, Mamba inherently preserves the intrinsic voxel-level spatial 
continuity that Transformers often sacrifice. However, while Mamba provides an 
ideal backbone, its standard scanning mechanisms treat all spatial dimensions 
uniformly. This generic approach fails to explicitly account for the pronounced 
anisotropy typical of EM data, where the axial resolution is often much lower 
than that of the transverse plane. Consequently, the distinct spatial 
relationships in neuronal structures cannot be optimally modeled by existing 
scanning strategies, highlighting the critical need for a mechanism tailored to 
differentiate and dynamically balance transverse and axial features.

In this paper, we propose NeuroMamba, a novel framework designed for 
accurate neuron segmentation by synergistically integrating local morphological 
cues with global spatial contexts. Specifically, we first design a 
channel-gated Boundary Discriminative Feature Extractor (BDFE) to 
capture fine-grained local details. This module enhances boundary precision 
across multiple channels, ensuring that delicate neuronal membranes are not 
overlooked. Complementing this local feature modeling, we introduce the 
Spatial Continuous Feature Extractor (SCFE) to efficiently model 
long-range voxel dependencies. Within this module, we incorporate a 
resolution-aware scanning mechanism into the Visual Mamba architecture 
to specifically address the limitations of generic scanning strategies in 
handling anisotropy. By utilizing resolution priors, this mechanism dynamically 
balances the modeling of transverse and axial features, enabling adaptive 
global modeling without patch partitioning. Finally, these multi-perspective 
features are synergistically integrated via a Cross Feature Interaction 
(CFI) module using a cross-modulation mechanism, yielding robust and accurate 
affinity predictions.

The main contributions of this paper are summarized as follows:
\begin{itemize}
	\item We propose NeuroMamba, a novel multi-perspective framework that 
	synergizes the linear complexity of Mamba for patch-free global modeling 
	with complementary local feature modeling. This architecture effectively 
	overcomes the receptive field limitations of CNNs and the boundary 
	imprecision issues of patch-based Transformers.
	
	\item We design two specialized feature extractors and a feature 
	interaction module: the BDFE enhances fine-grained morphological details, 
	the SCFE models long-range voxel dependencies, and the CFI dynamically 
	fuses these complementary features via cross-modulation.
	
	\item We introduce a novel resolution-aware scanning mechanism to address 
	the inflexibility of generic scanning strategies. By incorporating 
	resolution priors, this mechanism dynamically balances the modeling of 
	transverse and axial features to adaptively model spatial dependencies 
	across varying data resolutions, ranging from isotropic to highly 
	anisotropic.
	
	\item Our method achieves state-of-the-art performance on four public EM 
	datasets, demonstrating exceptional adaptability to both anisotropic and 
	isotropic resolutions.
\end{itemize}

\section{Related Work}

\subsection{Neuron Segmentation}
Deep learning-based approaches for neuron segmentation are broadly categorized 
into object-tracking-based and boundary-detection-based methods. 
Object-tracking-based methods~\cite{ffn,cross,schmidt2024roboem} reconstruct 
neurons individually, a process that is often computationally intensive. In 
contrast, boundary-detection-based methods have become the predominant 
paradigm; they first predict voxel affinities and then apply post-processing 
techniques~\cite{multicut,deepmulticut} to segment neurons, offering a more 
computationally efficient alternative.

Within this paradigm, CNNs have served as the foundational 
architecture~\cite{superhuman,mala,lsd,xiao2022deep,embedded}. Despite 
significant 
advances, these models are inherently constrained by their local receptive 
fields, which limits their ability to capture the long-range dependencies 
required to maintain the continuity of tortuous neuronal processes.

To address this core limitation of locality, architectural paradigms that 
integrate Vision Transformers with U-shaped networks, exemplified by models 
such as UNETR~\cite{UNETR} and SwinUNETR~\cite{SwinUNETR}, have been introduced 
to the 
field~\cite{self,appearance}. By modeling global dependencies between image 
patches, they 
can better preserve structural continuity. However, this patch-based approach 
presents inherent limitations. It often neglects fine-grained, 
voxel-level correlations within each patch, thereby compromising boundary 
precision. Furthermore, these models typically require extensive pre-training 
on large-scale datasets and exhibit limited adaptability to the anisotropic 
characteristics of EM data~\cite{self}. These complementary weaknesses of CNNs 
and Transformers highlight a clear need for a new architecture that can 
efficiently and holistically model both local details and global context.

\subsection{Visual Mamba}
Mamba~\cite{mamba}, built upon state space models (SSMs)~\cite{ssm}, has 
recently emerged as a compelling successor to Transformers for long-sequence 
modeling. It achieves linear complexity while demonstrating exceptional 
capability in capturing long-range 
dependencies, which has facilitated its rapid adoption in computer 
vision~\cite{visionmamba,localmamba,efficientvmamba,plainmamba} and medical 
image segmentation. Initial works like U-Mamba~\cite{U-Mamba} and 
SegMamba~\cite{SegMamba} successfully integrated Mamba blocks into U-Net 
architectures, demonstrating promising results on various 2D and 3D medical 
imaging tasks.

The application of Mamba has also been explored in the challenging domain of EM 
image segmentation. ViM-UNet~\cite{vimunet} was the first to adapt Mamba for 2D 
EM segmentation, while EMmamba~\cite{EMmamba} adopted the TSMamba block 
from SegMamba, leveraging pre-training to tackle the task of 3D EM 
segmentation. However, a critical limitation of these pioneering methods is 
their reliance on generic scanning mechanisms. As a core component, the 
scanning mechanism not only enhances computational efficiency but also encodes 
vital task-specific spatial information. Existing approaches either adopt 
scanning strategies from 2D applications (e.g., bidirectional 
scan~\cite{visionmamba}), video processing (3D bidirectional 
scan~\cite{videomamba}), or other domains like object detection (e.g., Hilbert 
scan~\cite{voxelmamba,mambaAD}). 

These generalized scanning strategies are suboptimal for 3D neuron segmentation 
because they do not explicitly account for the pronounced anisotropy present in 
most EM volumes. The complex, intertwining morphology of neurons requires a 
more nuanced approach to modeling spatial relationships. Consequently, the 
development of scanning mechanisms specifically tailored to differentiate and 
dynamically balance the modeling of transverse and axial features in EM data 
remains a critical and unexplored research direction.

\begin{figure}[t]
	\centering
	\includegraphics[width=1.0\columnwidth]{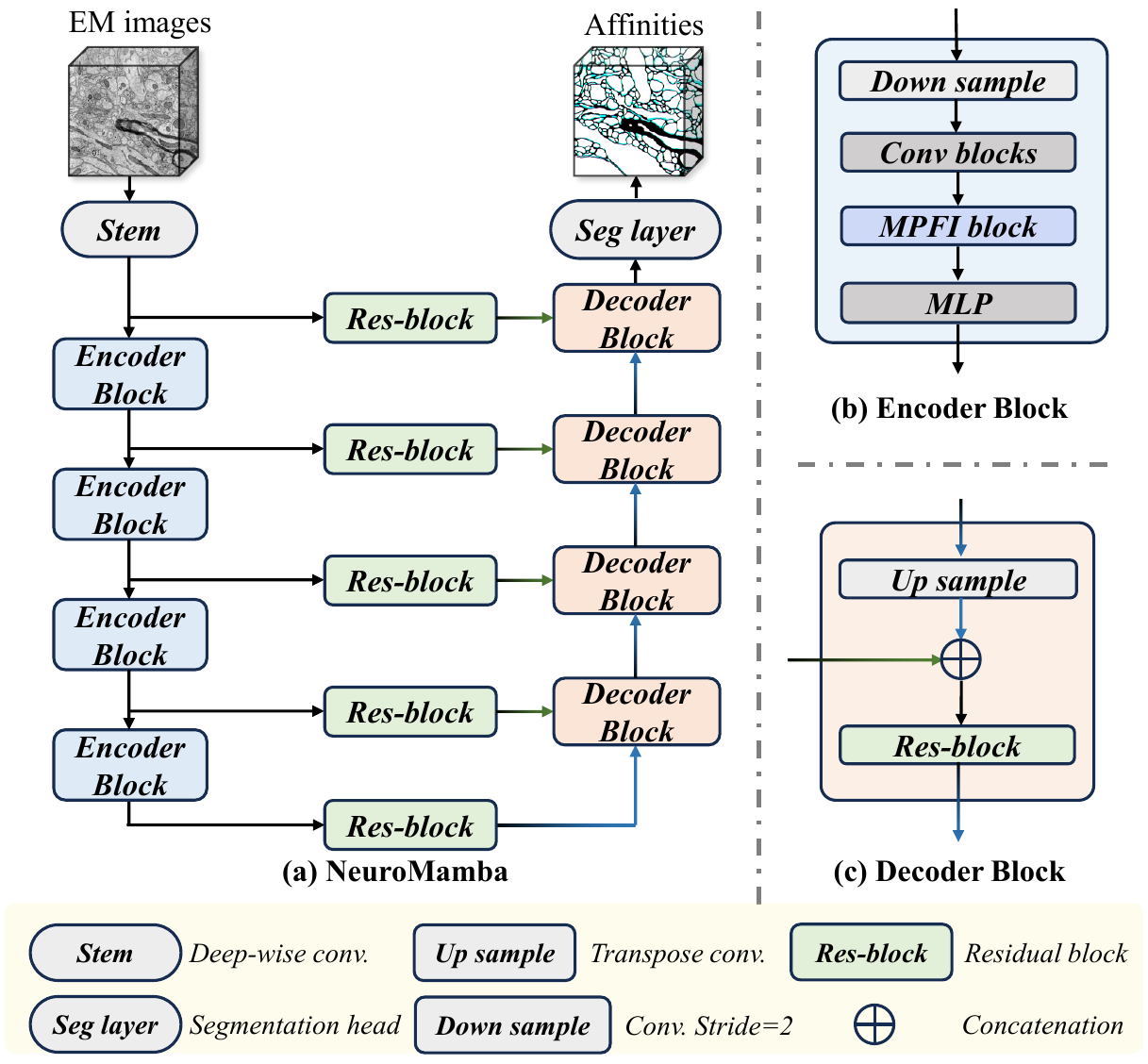} 
	\caption{Architecture of the proposed NeuroMamba. 
		(a) The overall pipeline of NeuroMamba. 
		(b) Details of the Encoder Block, where we introduce the MPFI block for 
		multi-perspective modeling. 
		(c) Details of the Decoder Block.}
	\label{fig2}
	\vspace{-1.0em}
\end{figure}

\section{Method}
\begin{figure*}[t]
	\centering
	\includegraphics[width=1.0\textwidth]{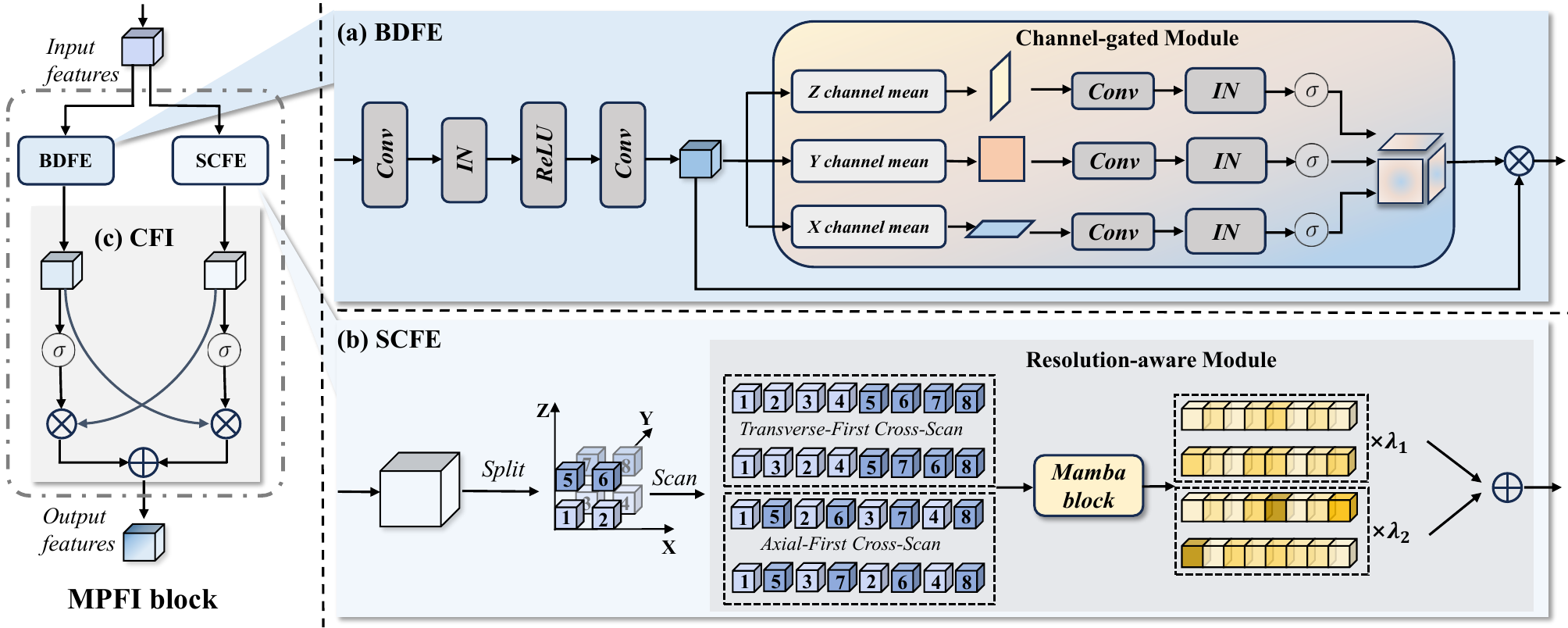} 
	\caption{Illustration of the Multi-Perspective Feature Interaction 
		(MPFI) block. The MPFI block includes two feature extractors and a 
		feature 
		interaction module: (a) Boundary Discriminative Feature Extractor 
		(BDFE), 
		(b) Spatial Continuous Feature Extractor (SCFE), and (c) Cross Feature 
		Interaction (CFI). Our MPFI block models neuronal affinity information 
		from 
		both local boundary discriminative features and global spatial 
		continuity 
		features.}
	\label{fig3}
\end{figure*}

In this section, we present the technical details of NeuroMamba. We first 
formally define the neuron segmentation task. Then, we introduce the three core 
components of our proposed Multi-Perspective Feature Interaction (MPFI) block, 
as illustrated in Fig.~\ref{fig3}: 
(1) the Boundary Discriminative Feature Extractor (BDFE), designed to 
capture fine-grained local boundary cues; 
(2) the Spatial Continuous Feature Extractor (SCFE), engineered to 
model long-range global dependencies without patching; and 
(3) the Cross Feature Interaction (CFI) module, which synergistically 
fuses these complementary features.

\subsection{Problem Definition}
Given a 3D EM image volume $\mathbf{I} \in \mathbb{R}^{D\times H\times W}$, 
the 
task of neuron segmentation is to predict the identity of each voxel, denoted 
as $\mathbf{S} \in \mathbb{N}^{D\times H\times W}$, where $D$, $H$, and $W$ 
represent the depth, height, and width of the volume, respectively. Each 
voxel's ID in $\mathbf{S}$ indicates the specific neuron to which it belongs, 
with ID 0 representing the background. We follow a two-stage boundary 
detection 
approach to predict the affinity $\mathbf{A} \in {[0,1]}^{3\times D\times 
	H\times W}$, which represents the probability that adjacent voxels in the 
	three 
directions belong to the same neuron. Subsequently, a post-processing step is 
applied to the affinity $\mathbf{A}$ to obtain the final neuron instance 
segmentation 
result $\mathbf{S}$.

\subsection{BDFE:Boundary Discriminative Feature Extractor}
To improve the accuracy of neuron boundary segmentation, especially in densely 
packed regions, we introduce a channel-gated module designed to enhance the 
extraction of neuronal morphological features. This module improves the model's 
ability to capture voxel affinities in fine-grained structures, thereby 
reducing merge and split errors in the final segmentation. A key challenge in 
this task stems from the elongated morphology of neurons. For such long-range, 
anisotropic structures, the large, square pooling windows employed in 
conventional 
CNNs~\cite{superhuman, mala} often incorporate extraneous information from 
adjacent yet unrelated regions, which can dilute the precision of boundary 
features~\cite{i:1}. 
Therefore, inspired by the success of strip pooling, we employ it within our 
channel-gated module to specifically address this issue.

Specifically, we pass the input $\mathbf{I} \in \mathbb{R}^{D\times H\times W}$ 
through a series of simple 
convolutions, normalizations, and activation functions to obtain features 
$\mathbf{I'} \in \mathbb{R}^{D\times H\times W}$, 
which are then fed into the channel-gated module. The first step of the 
channel-gated module involves performing strip convolutions along the $z$, $y$, 
and $x$ channels separately. The outputs $\mathbf{y^d} \in \mathbb{R}^D$, 
$\mathbf{y^h} \in \mathbb{R}^H$, and $\mathbf{y^w} \in \mathbb{R}^W$ for each 
corresponding channel can be represented as follows:
\begin{align}
	&\mathbf{I'} = Conv(ReLU(IN(Conv(\mathbf{I})))) \\
	&\mathbf{y^d_i} = \frac{1}{H\times W}\sum_{0\le j < H,0\le k < 
		W}^{}\mathbf{I'_{i,j,k}} \\
	&\mathbf{y^h_j} = \frac{1}{D\times W}\sum_{0\le i < D,0\le k < 
		W}^{}\mathbf{I'_{i,j,k}} \\
	&\mathbf{y^w_k} = \frac{1}{D\times H}\sum_{0\le i < D,0\le j < 
		H}^{}\mathbf{I'_{i,j,k}}
\end{align}

Unlike square pooling kernels, strip pooling in three dimensions generates 
long, narrow one-dimensional features that align with the elongated 
morphological characteristics of neuronal axons. These three features, 
$\mathbf{y^d}$, $\mathbf{y^h}$, and $\mathbf{y^w}$, encapsulate precise 
boundary information of neurons in three dimensions. Next, we apply 1D 
convolutions to interact with the three feature vectors, enhancing the 
representation of boundary information. The features are then normalized using 
an Instance Normalization (IN) layer and activated by a nonlinear activation 
function to obtain boundary position features for each dimension.
\begin{align}
	&\mathbf{z^d} = \sigma (IN(Conv^1(\mathbf{y^d}))) \\
	&\mathbf{z^h} = \sigma (IN(Conv^1(\mathbf{y^h}))) \\
	&\mathbf{z^w} = \sigma (IN(Conv^1(\mathbf{y^w}))) 
\end{align}

In the description above, $\sigma$ represents the sigmoid activation function, 
and $Conv^1$ denotes the one-dimensional convolution operation, where the 
parameters are shared across the three channels. The boundary position 
activation features 
$\mathbf{z^d}$, $\mathbf{z^h}$, and $\mathbf{z^w}$ are multiplied by the 
original features $\mathbf{I'}$, using a gating mechanism to capture the 
regions of interest. Finally, we obtain the output of the BDFE, denoted as 
$X_{local}$, as shown below:
\begin{equation}
	X_{local} = \mathbf{I'} \times \mathbf{z^d} \times \mathbf{z^h} 
	\times \mathbf{z^w}
	\label{eq8}
\end{equation}

\subsection{SCFE:Spatial Continuous Feature Extractor}
As established in previous sections, a core challenge in neuron segmentation is 
to model long-range spatial dependencies efficiently. While CNNs are limited by 
their local receptive fields, Transformer-based approaches, despite their 
global view, disrupt intrinsic spatial continuity by partitioning the data into 
patches. Furthermore, both architectures often exhibit inconsistent performance 
across EM datasets with varying degrees of anisotropy. This 
necessitates a novel architectural component capable of holistically capturing 
global 
voxel-level relationships without patching while dynamically adapting to 
varying 
data resolutions.

To this end, we base the SCFE on the Mamba architecture. 
The notable success of Mamba in its foundational domain—where, for instance, 
the Mamba-3B model has been shown to outperform Transformer-based models of 
equivalent size on language tasks~\cite{mamba}—motivates our investigation 
into its potential for neuronal reconstruction. A key advantage of Mamba is its 
linear complexity with respect to sequence length, which allows us to flatten 
an entire volumetric block into a single, long 1D sequence for processing. 
This obviates the need for patch partitioning—a strategy Transformers must 
employ to manage their quadratic complexity—thereby preserving intrinsic 
voxel-level spatial continuity.

However, while Mamba provides an ideal backbone, its standard scanning 
mechanisms are not inherently designed to handle the pronounced anisotropy of 
EM data. To address this limitation, we propose a novel resolution-aware 
module. 
This module employs a pair of specialized scanning mechanisms: a 
transverse-first cross-scan and an axial-first cross-scan, as 
illustrated in Fig.~\ref{fig3}. This approach is explicitly designed to 
differentiate between the high-resolution transverse planes (x-y) and the 
typically lower-resolution axial direction (z), allowing for a more nuanced 
modeling of anisotropic neuronal structures.

By design, this resolution-aware module dynamically adapts to varying data 
resolutions by leveraging the EM data's resolution as a prior. This prior 
enables the 
module to adjust the weights, $\lambda_1$ and $\lambda_2$, for the 
transverse-first and axial-first scans, respectively. Consequently, with 
increasing anisotropy in the neuronal volume, the SCFE block assigns a greater 
weight to the axial-first cross-scan ($\lambda_2$) and a lesser weight to the 
transverse-first cross-scan ($\lambda_1$). This enhances the model's ability to 
capture the continuity of axial data. Therefore, we establish the following 
relationship between $\lambda_1$ and $\lambda_2$:
\begin{align}
	&\lambda_1 + \lambda_2 = 2 \label{eq:9} \\
	&\lambda_2 = \alpha D_{ani} + \beta \label{eq:10}
\end{align}

Here, $\alpha$ and $\beta$ are hyperparameters. $D_{ani}$ represents 
the degree of anisotropy in the EM data, which can be computed using the axial 
resolution $R_a$ and transverse resolution $R_t$ as follows:
\begin{equation}
	D_{ani}=\frac{R_a}{R_t}
\end{equation}

Specifically, we flatten the 3D input features using the transverse-first 
cross-directional scanning mechanism to obtain $\mathbf{z^t_1}$ and 
$\mathbf{z^t_2}$, and the axial-first cross-directional scanning mechanism to 
obtain $\mathbf{z^a_1}$ and $\mathbf{z^a_2}$, where $t$ denotes transverse and 
$a$ denotes axial. These 
four sequences are then processed by the Mamba layer (denoted as $M$) to model 
global information. Subsequently, feature interactions are performed to obtain 
the SCFE output, denoted as $X_{global}$.
\begin{equation}
	X_{global}=\lambda_1 M(\mathbf{z^t_1})+\lambda_1 
	M(\mathbf{z^t_2})+\lambda_2 M(\mathbf{z^a_1})+\lambda_2 M(\mathbf{z^a_1})
	\label{eq9}
\end{equation}

\subsection{CFI:Cross Feature Interaction}
EM volumes typically contain neuron instances that span a wide range of scales, 
from large, continuous structures to small, intricate fragments. The local 
boundary features from the BDFE are essential for delineating these smaller 
instances, while the global continuity features from the SCFE are crucial for 
maintaining the integrity of larger ones. To address the challenge of 
effectively integrating these two perspectives, we introduce a dynamic feature 
interaction mechanism. In contrast to direct, addition-based fusion methods, 
our approach employs cross-modulation, which enables the model to adaptively 
process neurons of varying scales by dynamically modulating the interplay 
between local and global feature maps.

This interaction can be conceptualized as a reciprocal enhancement 
process, as illustrated in Fig.~\ref{fig3}(c). The global continuity 
map ($X_{global}$) serves as a spatial attention mechanism for the local 
boundary features ($X_{local}$), thereby enhancing feature salience in regions 
corresponding to large, continuous structures. Conversely, the local boundary 
map provides high-frequency details to refine the global features, thereby 
improving the precision of affinity predictions in structurally complex regions.

Specifically, given the local features $X_{local}$ from Eq.~\eqref{eq8} and the 
global features $X_{global}$ from Eq.~\eqref{eq9}, a reciprocal modulation is 
performed. This modulation is implemented by using each feature map to gate 
the other via a sigmoid function ($\sigma$), where $\odot$ denotes the 
element-wise Hadamard product:
\begin{align}
	&X'_{local}=X_{local}\odot \sigma(X_{global})=X_{local}\odot 
	\frac{1}{1+e^{-X_{global}}} \\
	&X'_{global}=X_{global}\odot \sigma(X_{local})=X_{global}\odot 
	\frac{1}{1+e^{-X_{local}}}
\end{align}
Following cross-modulation, the resulting representations $X'_{local}$ and 
$X'_{global}$ each integrate information from the complementary perspective. 
These modulated representations are then summed to produce the final output of 
the MPFI block, $O_{MPFI}$, which is synergistically enriched from both 
perspectives:
\begin{equation}
	O_{MPFI} = X'_{local} + X'_{global}
\end{equation}
This synergy enriches the final feature representation from both local and 
global perspectives, yielding more robust and accurate affinity predictions 
across multiple scales.

\begin{table*}[t]
	\centering
	\small
	\renewcommand{\arraystretch}{1.0}
	\setlength{\tabcolsep}{9.65pt}
	\begin{tabular}{cc|cccc|cccc}
		\toprule
		\multicolumn{2}{c|}{\multirow{2}{*}{Methods}} & 
		\multicolumn{4}{c|}{Waterz~\cite{mala}} & 
		\multicolumn{4}{c}{Multicut~\cite{multicut}} \\ 
		\cmidrule{3-10} 
		\multicolumn{2}{c|}{}                                                   
		& $\mathrm{VI}_s\downarrow$  & $\mathrm{VI}_m\downarrow$
		& $\mathrm{VI}\downarrow$    & $\mathrm{ARAND}\downarrow$          
		& $\mathrm{VI}_s\downarrow$  & $\mathrm{VI}_m\downarrow$  
		& $\mathrm{VI}\downarrow$    & $\mathrm{ARAND}\downarrow$ \\ 
		\midrule
		\multicolumn{1}{c}{\multirow{9}{*}{\rotatebox{90}{AC3/AC4}}}     
		& \multicolumn{1}{|c|}{\cellcolor{green_tab}MALA\cite{mala}}  
		& 0.892 & 0.331 
		& 1.224 \cellcolor[HTML]{EFEFEF}  
		& 0.135 \cellcolor[HTML]{EFEFEF}       
		& 0.827 & 0.406 
		& 1.233 \cellcolor[HTML]{EFEFEF}         
		& 0.131 \cellcolor[HTML]{EFEFEF} \\ 
		
		& 
		\multicolumn{1}{|c|}{\cellcolor{green_tab}Superhuman\cite{superhuman}} 
		
		& 0.770 & 0.289 
		& \underline{1.059} \cellcolor[HTML]{EFEFEF}  
		& \underline{0.108} \cellcolor[HTML]{EFEFEF}       
		& 0.794 & 0.316 
		& 1.110 \cellcolor[HTML]{EFEFEF}         
		& 0.105 \cellcolor[HTML]{EFEFEF} \\ 
		
		& \multicolumn{1}{|c|}{\cellcolor{green_tab}PEA\cite{embedded}}  
		& 0.812 & 0.285 
		& 1.097 \cellcolor[HTML]{EFEFEF}  
		& 0.118 \cellcolor[HTML]{EFEFEF}       
		& 0.728 & 0.396 
		& 1.124 \cellcolor[HTML]{EFEFEF}         
		& 0.122 \cellcolor[HTML]{EFEFEF} \\
		
		& \multicolumn{1}{|c|}{\cellcolor{green_tab}LSD\cite{lsd}}  
		& 0.923 & 0.320 
		& 1.246 \cellcolor[HTML]{EFEFEF}  
		& 0.122 \cellcolor[HTML]{EFEFEF}       
		& 0.982 & 0.348 
		& 1.330 \cellcolor[HTML]{EFEFEF}         
		& 0.128 \cellcolor[HTML]{EFEFEF} \\
		
		& \multicolumn{1}{|c|}{\cellcolor{oran_tab}UNETR\cite{UNETR}}
		& 0.952 & 0.812 
		& 1.765 \cellcolor[HTML]{EFEFEF}  
		& 0.334 \cellcolor[HTML]{EFEFEF}       
		& 0.931 & 0.992 
		& 1.922 \cellcolor[HTML]{EFEFEF}         
		& 0.373 \cellcolor[HTML]{EFEFEF} \\ 
		
		& \multicolumn{1}{|c|}{\cellcolor{oran_tab}SwinUNETR\cite{SwinUNETR}}
		& 0.873 & 0.490 
		& 1.364 \cellcolor[HTML]{EFEFEF}  
		& 0.238 \cellcolor[HTML]{EFEFEF}       
		& 0.841 & 0.425 
		& 1.266 \cellcolor[HTML]{EFEFEF}         
		& 0.171 \cellcolor[HTML]{EFEFEF} \\ 
		
		& \multicolumn{1}{|c|}{\cellcolor{blue_tab}U-Mamba\cite{U-Mamba}} 
		& 0.875 & 0.297 
		& 1.171 \cellcolor[HTML]{EFEFEF}  
		& 0.124 \cellcolor[HTML]{EFEFEF}       
		& 0.800 & 0.313 
		& 1.114 \cellcolor[HTML]{EFEFEF}         
		& 0.109 \cellcolor[HTML]{EFEFEF} \\ 
		
		& \multicolumn{1}{|c|}{\cellcolor{blue_tab}SegMamba\cite{SegMamba}}  
		& 0.801 & 0.287 
		& 1.088 \cellcolor[HTML]{EFEFEF}  
		& 0.109 \cellcolor[HTML]{EFEFEF}       
		& 0.796 & 0.286 
		& \underline{1.082} \cellcolor[HTML]{EFEFEF}         
		& \underline{0.102} \cellcolor[HTML]{EFEFEF} \\ 
		
		& \multicolumn{1}{|c|}{\cellcolor{blue_tab}EMmamba\cite{EMmamba}}  
		& 0.906 & 0.411 
		& 1.317 \cellcolor[HTML]{EFEFEF}  
		& 0.138 \cellcolor[HTML]{EFEFEF}       
		& 0.815 & 0.595 
		& 1.411 \cellcolor[HTML]{EFEFEF}         
		& 0.165 \cellcolor[HTML]{EFEFEF} \\ 
		
		& \multicolumn{1}{|c|}{\cellcolor{blue_tab}Ours}  
		& 0.752 & 0.269 
		& \textbf{1.020} \cellcolor[HTML]{EFEFEF}  
		& \textbf{0.104} \cellcolor[HTML]{EFEFEF}       
		& 0.712 & 0.259 
		& \textbf{0.971} \cellcolor[HTML]{EFEFEF}         
		& \textbf{0.090} \cellcolor[HTML]{EFEFEF} \\                      
		
		\midrule 
		\midrule
		
		\multicolumn{1}{c}{\multirow{9}{*}{\rotatebox{90}{CREMI-A}}}     
		& \multicolumn{1}{|c|}{\cellcolor{green_tab}MALA\cite{mala}}  
		& 0.509 & 0.475 
		& 0.984 \cellcolor[HTML]{EFEFEF}  
		& 0.204 \cellcolor[HTML]{EFEFEF}       
		& 0.505 & 0.437 
		& 0.942 \cellcolor[HTML]{EFEFEF}         
		& 0.179 \cellcolor[HTML]{EFEFEF} \\ 
		
		& 
		\multicolumn{1}{|c|}{\cellcolor{green_tab}Superhuman\cite{superhuman}} 
		& 0.515 & 0.463 
		& 0.978 \cellcolor[HTML]{EFEFEF}  
		& 0.190 \cellcolor[HTML]{EFEFEF}       
		& 0.517 & 0.432 
		& 0.949 \cellcolor[HTML]{EFEFEF}         
		& 0.176 \cellcolor[HTML]{EFEFEF} \\ 
		
		& \multicolumn{1}{|c|}{\cellcolor{green_tab}PEA\cite{embedded}}  
		& 0.530 & 0.390 
		& \underline{0.920} \cellcolor[HTML]{EFEFEF}  
		& 0.186 \cellcolor[HTML]{EFEFEF}       
		& 0.557 & 0.378 
		& \underline{0.935} \cellcolor[HTML]{EFEFEF}         
		& \underline{0.175} \cellcolor[HTML]{EFEFEF} \\
		
		& \multicolumn{1}{|c|}{\cellcolor{green_tab}LSD\cite{lsd}}  
		& 0.489 & 0.440 
		& 0.929 \cellcolor[HTML]{EFEFEF}  
		& \underline{0.183} \cellcolor[HTML]{EFEFEF}       
		& 0.569 & 0.466 
		& 1.035 \cellcolor[HTML]{EFEFEF}         
		& 0.217 \cellcolor[HTML]{EFEFEF} \\
		
		& \multicolumn{1}{|c|}{\cellcolor{oran_tab}UNETR\cite{UNETR}} 
		& 0.777 & 1.027 
		& 1.805 \cellcolor[HTML]{EFEFEF}  
		& 0.444 \cellcolor[HTML]{EFEFEF}       
		& 0.558 & 0.728 
		& 1.286 \cellcolor[HTML]{EFEFEF}         
		& 0.314 \cellcolor[HTML]{EFEFEF} \\ 
		
		& \multicolumn{1}{|c|}{\cellcolor{oran_tab}SwinUNETR\cite{SwinUNETR}}  
		& 0.721 & 0.819 
		& 1.540 \cellcolor[HTML]{EFEFEF}  
		& 0.365 \cellcolor[HTML]{EFEFEF}       
		& 0.540 & 0.525 
		& 1.066 \cellcolor[HTML]{EFEFEF}         
		& 0.224 \cellcolor[HTML]{EFEFEF} \\ 
		
		& \multicolumn{1}{|c|}{\cellcolor{blue_tab}U-Mamba\cite{U-Mamba}}  
		& 0.565 & 0.509 
		& 1.074 \cellcolor[HTML]{EFEFEF}  
		& 0.224 \cellcolor[HTML]{EFEFEF}       
		& 0.521 & 0.482 
		& 1.003 \cellcolor[HTML]{EFEFEF}         
		& 0.201 \cellcolor[HTML]{EFEFEF} \\ 
		
		& \multicolumn{1}{|c|}{\cellcolor{blue_tab}SegMamba\cite{SegMamba}} 
		& 0.565 & 0.566 
		& 1.131 \cellcolor[HTML]{EFEFEF}  
		& 0.247 \cellcolor[HTML]{EFEFEF}       
		& 0.513 & 0.494 
		& 1.007 \cellcolor[HTML]{EFEFEF}         
		& 0.201 \cellcolor[HTML]{EFEFEF} \\ 
		
		& \multicolumn{1}{|c|}{\cellcolor{blue_tab}EMmamba\cite{EMmamba}}  
		& 0.610 & 0.693 
		& 1.303 \cellcolor[HTML]{EFEFEF}  
		& 0.287 \cellcolor[HTML]{EFEFEF}       
		& 0.529 & 0.594 
		& 1.122 \cellcolor[HTML]{EFEFEF}         
		& 0.277 \cellcolor[HTML]{EFEFEF} \\
		
		& \multicolumn{1}{|c|}{\cellcolor{blue_tab}Ours} 
		& 0.477 & 0.388 
		& \textbf{0.865} \cellcolor[HTML]{EFEFEF}  
		& \textbf{0.142} \cellcolor[HTML]{EFEFEF}       
		& 0.484 & 0.370 
		& \textbf{0.853} \cellcolor[HTML]{EFEFEF}         
		& \textbf{0.137} \cellcolor[HTML]{EFEFEF} \\                    
		
		\midrule 
		\midrule
		
		\multicolumn{1}{c}{\multirow{9}{*}{\rotatebox{90}{CREMI-B}}}     
		& \multicolumn{1}{|c|}{\cellcolor{green_tab}MALA\cite{mala}} 
		& 0.931 & 0.655 
		& 1.586 \cellcolor[HTML]{EFEFEF}  
		& 0.170 \cellcolor[HTML]{EFEFEF}       
		& 1.039 & 0.351 
		& 1.391 \cellcolor[HTML]{EFEFEF}         
		& 0.088 \cellcolor[HTML]{EFEFEF} \\ 
		
		& 
		\multicolumn{1}{|c|}{\cellcolor{green_tab}Superhuman\cite{superhuman}}  
		& 0.696 & 0.873 
		& \underline{1.569} \cellcolor[HTML]{EFEFEF}  
		& 0.185 \cellcolor[HTML]{EFEFEF}       
		& 0.930 & 0.432 
		& \underline{1.362} \cellcolor[HTML]{EFEFEF}         
		& \underline{0.083} \cellcolor[HTML]{EFEFEF} \\ 
		
		& \multicolumn{1}{|c|}{\cellcolor{green_tab}PEA\cite{embedded}}  
		& 0.942 & 0.326 
		& \textbf{1.268} \cellcolor[HTML]{EFEFEF}  
		& \textbf{0.080} \cellcolor[HTML]{EFEFEF}       
		& 1.213 & 0.232 
		& 1.446 \cellcolor[HTML]{EFEFEF}         
		& 0.084 \cellcolor[HTML]{EFEFEF} \\
		
		& \multicolumn{1}{|c|}{\cellcolor{green_tab}LSD\cite{lsd}}  
		& 1.561 & 0.789 
		& 2.350 \cellcolor[HTML]{EFEFEF}  
		& 0.285 \cellcolor[HTML]{EFEFEF}       
		& 1.848 & 0.770 
		& 2.618 \cellcolor[HTML]{EFEFEF}         
		& 0.286 \cellcolor[HTML]{EFEFEF} \\
		
		& \multicolumn{1}{|c|}{\cellcolor{oran_tab}UNETR\cite{UNETR}} 
		& 2.337 & 3.091 
		& 5.428 \cellcolor[HTML]{EFEFEF}  
		& 0.763 \cellcolor[HTML]{EFEFEF}       
		& 2.303 & 0.821 
		& 3.124 \cellcolor[HTML]{EFEFEF}         
		& 0.357 \cellcolor[HTML]{EFEFEF} \\ 
		
		& \multicolumn{1}{|c|}{\cellcolor{oran_tab}SwinUNETR\cite{SwinUNETR}}  
		& 1.754 & 1.488 
		& 3.243 \cellcolor[HTML]{EFEFEF}  
		& 0.418 \cellcolor[HTML]{EFEFEF}       
		& 2.187 & 0.551 
		& 2.738 \cellcolor[HTML]{EFEFEF}         
		& 0.316 \cellcolor[HTML]{EFEFEF} \\ 
		
		& \multicolumn{1}{|c|}{\cellcolor{blue_tab}U-Mamba\cite{U-Mamba}}
		& 0.984 & 0.825 
		& 1.809 \cellcolor[HTML]{EFEFEF}  
		& 0.223 \cellcolor[HTML]{EFEFEF}       
		& 1.080 & 0.599 
		& 1.679 \cellcolor[HTML]{EFEFEF}         
		& 0.171 \cellcolor[HTML]{EFEFEF} \\ 
		
		& \multicolumn{1}{|c|}{\cellcolor{blue_tab}SegMamba\cite{SegMamba}} 
		& 1.036 & 1.202 
		& 2.239 \cellcolor[HTML]{EFEFEF}  
		& 0.322 \cellcolor[HTML]{EFEFEF}       
		& 1.126 & 0.612 
		& 1.738 \cellcolor[HTML]{EFEFEF}         
		& 0.147 \cellcolor[HTML]{EFEFEF} \\ 
		
		& \multicolumn{1}{|c|}{\cellcolor{blue_tab}EMmamba\cite{EMmamba}}  
		& 1.389 & 1.667 
		& 3.056 \cellcolor[HTML]{EFEFEF}  
		& 0.458 \cellcolor[HTML]{EFEFEF}       
		& 1.177 & 1.093 
		& 2.270 \cellcolor[HTML]{EFEFEF}         
		& 0.310 \cellcolor[HTML]{EFEFEF} \\
		
		& \multicolumn{1}{|c|}{\cellcolor{blue_tab}Ours} 
		& 0.713 & 0.556 
		& \textbf{1.268} \cellcolor[HTML]{EFEFEF}  
		& \underline{0.087} \cellcolor[HTML]{EFEFEF}       
		& 0.898 & 0.408 
		& \textbf{1.305} \cellcolor[HTML]{EFEFEF}         
		& \textbf{0.073} \cellcolor[HTML]{EFEFEF} \\		                   
		
		\midrule 
		\midrule
		
		\multicolumn{1}{c}{\multirow{9}{*}{\rotatebox{90}{CREMI-C}}}     
		& \multicolumn{1}{|c|}{\cellcolor{green_tab}MALA\cite{mala}} 
		& 1.026 & 0.635 
		& 1.661 \cellcolor[HTML]{EFEFEF}  
		& 0.195 \cellcolor[HTML]{EFEFEF}       
		& 0.967 & 0.560 
		& 1.527 \cellcolor[HTML]{EFEFEF}         
		& 0.162 \cellcolor[HTML]{EFEFEF} \\ 
		
		& 
		\multicolumn{1}{|c|}{\cellcolor{green_tab}Superhuman\cite{superhuman}}  
		& 0.998 & 0.475 
		& \underline{1.473} \cellcolor[HTML]{EFEFEF}  
		& 0.142 \cellcolor[HTML]{EFEFEF}       
		& 0.951 & 0.516 
		& 1.467 \cellcolor[HTML]{EFEFEF}         
		& \underline{0.135} \cellcolor[HTML]{EFEFEF} \\ 
		
		& \multicolumn{1}{|c|}{\cellcolor{green_tab}PEA\cite{embedded}}  
		& 1.299 & 0.252 
		& 1.551 \cellcolor[HTML]{EFEFEF}  
		& \underline{0.135} \cellcolor[HTML]{EFEFEF}       
		& 1.066 & 0.360 
		& \textbf{1.426} \cellcolor[HTML]{EFEFEF}         
		& 0.145 \cellcolor[HTML]{EFEFEF} \\
		
		& \multicolumn{1}{|c|}{\cellcolor{green_tab}LSD\cite{lsd}}  
		& 1.095 & 0.583 
		& 1.678 \cellcolor[HTML]{EFEFEF}  
		& 0.190 \cellcolor[HTML]{EFEFEF}       
		& 1.189 & 0.694 
		& 1.883 \cellcolor[HTML]{EFEFEF}         
		& 0.218 \cellcolor[HTML]{EFEFEF} \\
		
		& \multicolumn{1}{|c|}{\cellcolor{oran_tab}UNETR\cite{UNETR}}  
		& 1.388 & 1.412 
		& 2.800 \cellcolor[HTML]{EFEFEF}  
		& 0.358 \cellcolor[HTML]{EFEFEF}       
		& 1.238 & 0.774 
		& 2.012 \cellcolor[HTML]{EFEFEF}         
		& 0.257 \cellcolor[HTML]{EFEFEF} \\ 
		
		& \multicolumn{1}{|c|}{\cellcolor{oran_tab}SwinUNETR\cite{SwinUNETR}}  
		& 1.267 & 1.021 
		& 2.288 \cellcolor[HTML]{EFEFEF}  
		& 0.297 \cellcolor[HTML]{EFEFEF}       
		& 1.262 & 0.748 
		& 2.010 \cellcolor[HTML]{EFEFEF}         
		& 0.255 \cellcolor[HTML]{EFEFEF} \\ 
		
		& \multicolumn{1}{|c|}{\cellcolor{blue_tab}U-Mamba\cite{U-Mamba}}
		& 1.032 & 0.491 
		& 1.523 \cellcolor[HTML]{EFEFEF}  
		& 0.161 \cellcolor[HTML]{EFEFEF}       
		& 0.994 & 0.522 
		& 1.516 \cellcolor[HTML]{EFEFEF}         
		& 0.151 \cellcolor[HTML]{EFEFEF} \\ 
		
		& \multicolumn{1}{|c|}{\cellcolor{blue_tab}SegMamba\cite{SegMamba}}  
		& 1.030 & 0.502 
		& 1.532 \cellcolor[HTML]{EFEFEF}  
		& 0.158 \cellcolor[HTML]{EFEFEF}       
		& 1.016 & 0.514 
		& 1.530 \cellcolor[HTML]{EFEFEF}         
		& 0.172 \cellcolor[HTML]{EFEFEF} \\ 
		
		& \multicolumn{1}{|c|}{\cellcolor{blue_tab}EMmamba\cite{EMmamba}}  
		& 1.326 & 0.764 
		& 2.090 \cellcolor[HTML]{EFEFEF}  
		& 0.219 \cellcolor[HTML]{EFEFEF}       
		& 1.096 & 1.173 
		& 2.269 \cellcolor[HTML]{EFEFEF}         
		& 0.263 \cellcolor[HTML]{EFEFEF} \\
		
		& \multicolumn{1}{|c|}{\cellcolor{blue_tab}Ours}  
		& 0.930 & 0.452 
		& \textbf{1.382} \cellcolor[HTML]{EFEFEF}  
		& \textbf{0.128} \cellcolor[HTML]{EFEFEF}       
		& 0.938 & 0.492 
		& \underline{1.430} \cellcolor[HTML]{EFEFEF}         
		& \textbf{0.129} \cellcolor[HTML]{EFEFEF} \\
		
		\bottomrule
	\end{tabular}
	\caption{Quantitative comparisons of different methods on the AC3/AC4 and 
		CREMI datasets. \protect\sethlcolor{green_tab}\hl{Green}, 
		\protect\sethlcolor{oran_tab}\hl{orange} and 
		\protect\sethlcolor{blue_tab}\hl{blue} backgrounds indicate CNN-based, 
		Transformer-based, and Mamba-based methods, respectively. \textbf{Bold} 
		and \underline{underlined} items indicate the first and second scoring 
		results.}
	\label{tab1}
\end{table*}

\begin{table}[t]
	\centering
	\small
	\renewcommand{\arraystretch}{1.0}
	\setlength{\tabcolsep}{3.8pt}
	\begin{tabular}{cc|cccc}
		\toprule
		\multicolumn{2}{c|}{\multirow{2}{*}{Methods}} &  
		\multicolumn{4}{c}{Multicut~\cite{multicut}} \\ 
		\cmidrule{3-6} 
		\multicolumn{2}{c|}{}                                                   
		& $\mathrm{VI}_s\downarrow$  & $\mathrm{VI}_m\downarrow$
		& $\mathrm{VI}\downarrow$    & $\mathrm{ARAND}\downarrow$ \\ 
		\midrule
		\multicolumn{1}{c}{\multirow{9}{*}{\rotatebox{90}{FIB25}}}     
		& \multicolumn{1}{|c|}{\cellcolor{green_tab}MALA\cite{mala}}  
		& 1.098 & 1.056 
		& 2.154 \cellcolor[HTML]{EFEFEF}  
		& 0.216 \cellcolor[HTML]{EFEFEF} \\ 
		
		& 
		\multicolumn{1}{|c|}{\cellcolor{green_tab}Superhuman\cite{superhuman}}  
		& 1.071 & 0.888 
		& 1.959 \cellcolor[HTML]{EFEFEF}  
		& 0.166 \cellcolor[HTML]{EFEFEF} \\ 
		
		& \multicolumn{1}{|c|}{\cellcolor{green_tab}PEA\cite{embedded}}  
		& 0.950 & 1.017 
		& 1.967 \cellcolor[HTML]{EFEFEF}  
		& 0.200 \cellcolor[HTML]{EFEFEF} \\
		
		& \multicolumn{1}{|c|}{\cellcolor{green_tab}LSD\cite{lsd}}  
		& 1.217 & 0.821 
		& 2.038 \cellcolor[HTML]{EFEFEF}  
		& 0.181 \cellcolor[HTML]{EFEFEF} \\ 
		
		& \multicolumn{1}{|c|}{\cellcolor{oran_tab}UNETR\cite{UNETR}}  
		& 1.008 & 0.915 
		& 1.923 \cellcolor[HTML]{EFEFEF}  
		& \underline{0.162} \cellcolor[HTML]{EFEFEF} \\ 
		
		& \multicolumn{1}{|c|}{\cellcolor{oran_tab}SwinUNETR\cite{SwinUNETR}}  
		& 1.014 & 0.901 
		& \underline{1.915} \cellcolor[HTML]{EFEFEF}  
		& 0.168 \cellcolor[HTML]{EFEFEF} \\ 
		
		& \multicolumn{1}{|c|}{\cellcolor{blue_tab}U-Mamba\cite{U-Mamba}}  
		& 1.058 & 0.897 
		& 1.955 \cellcolor[HTML]{EFEFEF}  
		& 0.173 \cellcolor[HTML]{EFEFEF} \\ 
		
		& \multicolumn{1}{|c|}{\cellcolor{blue_tab}SegMamba\cite{SegMamba}}  
		& 1.060 & 0.866 
		& 1.926 \cellcolor[HTML]{EFEFEF}  
		& 0.167 \cellcolor[HTML]{EFEFEF} \\ 
		
		& \multicolumn{1}{|c|}{\cellcolor{blue_tab}EMmamba\cite{EMmamba}}  
		& 1.068 & 0.926 
		& 1.994 \cellcolor[HTML]{EFEFEF}  
		& 0.177 \cellcolor[HTML]{EFEFEF} \\ 
		
		& \multicolumn{1}{|c|}{\cellcolor{blue_tab}Ours}  
		& 1.041 & 0.851 
		& \textbf{1.893} \cellcolor[HTML]{EFEFEF}  
		& \textbf{0.159} \cellcolor[HTML]{EFEFEF} \\
		
		\bottomrule
	\end{tabular}
	\caption{Quantitative comparisons of different methods on the FIB25 
		datasets. \textbf{Bold} and \underline{underlined} items indicate the 
		first and second scoring results.}
	\label{tab2}
\end{table}

\begin{table}[t]
	\centering
	\small
	\renewcommand{\arraystretch}{1.0}
	\setlength{\tabcolsep}{2.0pt}
	\begin{tabular}{cc|cc|cc}
		\toprule
		\multicolumn{2}{c|}{\multirow{2}{*}{Methods}} &  
		\multicolumn{2}{c|}{Waterz} &
		\multicolumn{2}{c}{Multicut} \\ 
		\cmidrule{3-6} 
		\multicolumn{2}{c|}{} &
		\multicolumn{2}{c|}{$\mathrm{VI}\downarrow$ \qquad 
			$\mathrm{ARAND}\downarrow$} &
		\multicolumn{2}{c}{$\mathrm{VI}\downarrow$ \qquad 
			$\mathrm{ARAND}\downarrow$} \\ 
		\midrule        
		\multicolumn{1}{c}{\multirow{6}{*}{\rotatebox{90}{Kasthuri}}} 
		& \multicolumn{1}{|c|}{\cellcolor{green_tab}Superhuman\cite{superhuman}}
		& \underline{0.840} & 0.320 & \underline{0.829} & \underline{0.325} \\ 
		& \multicolumn{1}{|c|}{\cellcolor{green_tab}PEA\cite{embedded}}
		& 0.843 & 0.324 & 1.098 & 0.350 \\
		& \multicolumn{1}{|c|}{\cellcolor{green_tab}LSD\cite{lsd}}
		& 0.860 & \textbf{0.298} & 1.213 & 0.489 \\
		& \multicolumn{1}{|c|}{\cellcolor{oran_tab}SwinUNETR\cite{SwinUNETR}}
		& 1.259 & 0.605 & 1.148 & 0.538 \\
		& \multicolumn{1}{|c|}{\cellcolor{blue_tab}SegMamba\cite{SegMamba}}
		& 0.910 & 0.436 & 0.974 & 0.512 \\
		& \multicolumn{1}{|c|}{\cellcolor{blue_tab}Ours}
		& \textbf{0.787} & \underline{0.303} & \textbf{0.761} & \textbf{0.301} 
		\\
		
		\bottomrule
	\end{tabular}
	\caption{Quantitative comparisons of different methods on the Kasthuri 
		datasets. \textbf{Bold} and \underline{underlined} items indicate 
		the first and second scoring results.}
	\label{tab22}
\end{table}

\begin{table}[t]
	\centering
	\small
	\setlength{\tabcolsep}{1pt}
	\begin{tabular}{c|cccc}
		\toprule
		\multicolumn{1}{c|}{Methods} 
		& Params(M) & FLOPs(G) & Latency(s)
		& $\mathrm{ARAND}\downarrow$ \\ 
		\midrule  
		
		\multicolumn{1}{c|}{\cellcolor{green_tab}MALA\cite{mala}}  
		& 84.0 & 413.6 & 0.027 & 0.179 \\ 
		
		\multicolumn{1}{c|}{\cellcolor{green_tab}Superhuman\cite{superhuman}}  
		& 1.6 & 185.9 & 0.046 & 0.176 \\ 
		
		\multicolumn{1}{c|}{\cellcolor{green_tab}PEA\cite{embedded}}  
		& 2.1 & 177.2 &0.044 & \underline{0.175} \\  
		
		\multicolumn{1}{c|}{\cellcolor{green_tab}LSD\cite{lsd}}  
		& 1.6 & 186.1 & 0.046 & 0.217 \\
		
		\multicolumn{1}{c|}{\cellcolor{oran_tab}UNETR\cite{UNETR}}  
		& 115.2 & 752.3 & 0.060 & 0.314 \\ 
		
		\multicolumn{1}{c|}{\cellcolor{oran_tab}SwinUNETR\cite{SwinUNETR}}  
		& 62.0 & 440.3 &0.068 & 0.224 \\ 
		
		\multicolumn{1}{c|}{\cellcolor{blue_tab}U-Mamba\cite{U-Mamba}}  
		& 3.51 & 620.1 & 0.063 & 0.201 \\     
		
		\multicolumn{1}{c|}{\cellcolor{blue_tab}SegMamba\cite{SegMamba}}  
		& 64.2 & 1213.9 &0.067 & 0.201 \\   
		
		\multicolumn{1}{c|}{\cellcolor{blue_tab}EMmamba\cite{EMmamba}}  
		& 26.8 & 323.3 &0.054 & 0.277\\   
		
		\multicolumn{1}{c|}{\cellcolor{blue_tab}Ours}  
		& 2.1 & 220.8 &0.059 & \textbf{0.137} \\
		
		\bottomrule
	\end{tabular}
	\caption{Efficiency comparison of different methods on CREMI-A.}
	\label{tab4}
\end{table}

\noindent\begin{table*}[tp]
	\centering
	\subfloat[Ablation of BDFE and SCFE.]{
		\label{tab3_2}
		\renewcommand{\arraystretch}{0.8}
		\setlength\tabcolsep{2pt}
		\resizebox{0.45\linewidth}{!}{
			\begin{tabular}{cc|cc|cc}
				\toprule
				\multicolumn{1}{c}{\multirow{2}{*}{BDFE}} &  
				\multicolumn{1}{c|}{\multirow{2}{*}{SCFE}} &
				\multicolumn{2}{c|}{Waterz} &
				\multicolumn{2}{c}{Multicut} \\ 
				\cmidrule{3-6} 
				\multicolumn{2}{c|}{} &
				\multicolumn{2}{c|}{$\mathrm{VI}\downarrow$ \quad 
					$\mathrm{ARAND}\downarrow$} &
				\multicolumn{2}{c}{$\mathrm{VI}\downarrow$ \quad 
					$\mathrm{ARAND}\downarrow$} \\ 
				\midrule     
				\multicolumn{1}{c}{}  & \multicolumn{1}{c|}{}  
				& 0.895 & 0.163 & 0.906 & 0.169 \\ 
				\multicolumn{1}{c}{\checkmark}  & \multicolumn{1}{c|}{} 
				& 0.889 & 0.158 & 0.867 & 0.145 \\  
				\multicolumn{1}{c}{}  & \multicolumn{1}{c|}{\checkmark}  
				& 0.892 & 0.158 & 0.886 & 0.152 \\ 
				\multicolumn{1}{c}{\checkmark}  & 
				\multicolumn{1}{c|}{\checkmark}  
				& \textbf{0.865} & \textbf{0.142} 
				& \textbf{0.853} & \textbf{0.137} \\ 
				\bottomrule
			\end{tabular}
		}
	} \hfill
	\subfloat[Ablation of CFI.]{
		\label{tab3_3}
		\renewcommand{\arraystretch}{0.98}
		\setlength\tabcolsep{2pt}
		\resizebox{0.45\linewidth}{!}{
			\begin{tabular}{cc|cc|cc}
				\toprule
				\multicolumn{2}{c|}{\multirow{2}{*}{Methods}} &  
				\multicolumn{2}{c|}{Waterz} &
				\multicolumn{2}{c}{Multicut} \\ 
				\cmidrule{3-6} 
				\multicolumn{2}{c|}{} &
				\multicolumn{2}{c|}{$\mathrm{VI}\downarrow$ \qquad 
					$\mathrm{ARAND}\downarrow$} &
				\multicolumn{2}{c}{$\mathrm{VI}\downarrow$ \qquad 
					$\mathrm{ARAND}\downarrow$} \\ 
				\midrule     
				\multicolumn{2}{c|}{addition}
				& 0.915 & 0.168 & \textbf{0.852} & 0.142 \\ 
				\multicolumn{2}{c|}{multiplication}  
				& 0.929 & 0.169 & 0.894 & 0.154 \\
				\multicolumn{2}{c|}{concatenation}  
				& 0.889 & 0.158 & 0.869 & 0.150 \\ 
				\multicolumn{2}{c|}{interaction} 
				& \textbf{0.865} & \textbf{0.142} & 0.853 & \textbf{0.137} \\
				\bottomrule
			\end{tabular}
		}
	} \hfill
	\subfloat[Ablation of resolution-aware module.]{
		\label{tab3_4}
		\renewcommand{\arraystretch}{0.3}
		\setlength\tabcolsep{2.0pt}
		\resizebox{0.31\linewidth}{!}{
			\begin{tabular}{cc|cccc}
				\toprule
				\multicolumn{2}{c|}{\multirow{2}{*}{Datasets}} &  
				\multicolumn{2}{c|}{$\mathrm{VI}\downarrow$} &
				\multicolumn{2}{c}{$\mathrm{ARAND}\downarrow$} \\ 
				\cmidrule{3-6} 
				\multicolumn{2}{c|}{} &
				\multicolumn{1}{c}{w/o Pri.} &
				\multicolumn{1}{c|}{w Pri.} &
				\multicolumn{1}{c}{w/o Pri.} &
				\multicolumn{1}{c}{w Pri.} \\ 
				\midrule   
				\multicolumn{2}{c|}{AC3/AC4}  
				& 1.002 & \multicolumn{1}{c|}{\textbf{0.971}} 
				& 0.101 & \textbf{0.090}  \\  
				\midrule
				\multicolumn{2}{c|}{CREMI-A}  
				& 0.859 & \multicolumn{1}{c|}{\textbf{0.853}} 
				& 0.140 & \textbf{0.137}  \\  
				\midrule
				\multicolumn{2}{c|}{FIB25}  
				& 1.926 & \multicolumn{1}{c|}{\textbf{1.893}}  
				& 0.161 & \textbf{0.159}  \\ 
				\bottomrule
			\end{tabular}
		}
	} \hfill
	\subfloat[Effectiveness of Mamba.]{
		\label{tab3_5}
		\renewcommand{\arraystretch}{1.0}
		\setlength\tabcolsep{1.0pt}
		\resizebox{0.24\linewidth}{!}{
			\begin{tabular}{c|cc}
				\toprule
				\multicolumn{1}{c|}{Methods} 
				& $\mathrm{VI}\downarrow$
				& $\mathrm{ARAND}\downarrow$ \\ 
				\midrule     	
				\multicolumn{1}{c|}{UNETR}  
				& 1.286 & 0.314 \\ 
				
				\multicolumn{1}{c|}{UNEM}  
				& 1.278 & 0.281 \\  
				
				\multicolumn{1}{c|}{Ours}  
				& \textbf{0.853} & \textbf{0.137} \\ 
				
				\bottomrule
			\end{tabular}
		}
	} \hfill
	\subfloat[Ablation of strip pooling.]{
		\label{tab3_6}
		\renewcommand{\arraystretch}{1.4}
		\setlength\tabcolsep{4.0pt}
		\resizebox{0.32\linewidth}{!}{
			\begin{tabular}{c|cc}
				\toprule
				\multicolumn{1}{c|}{Methods} 
				& $\mathrm{VI}\downarrow$
				& $\mathrm{ARAND}\downarrow$ \\ 
				\midrule     	
				\multicolumn{1}{c|}{Square Pooling}  
				& 0.860 & 0.146 \\ 
				
				\multicolumn{1}{c|}{Strip Pooling}  
				& \textbf{0.853} & \textbf{0.137} \\  
				
				\bottomrule
			\end{tabular}
		}
	} \hfill
	\caption{Ablation studies and comparison analysis.}
	\label{table:ablation_all}
\end{table*}

\section{Experiments}
\subsection{Datasets}
\begin{figure*}[t]
	\centering
	\includegraphics[width=1.0\textwidth]{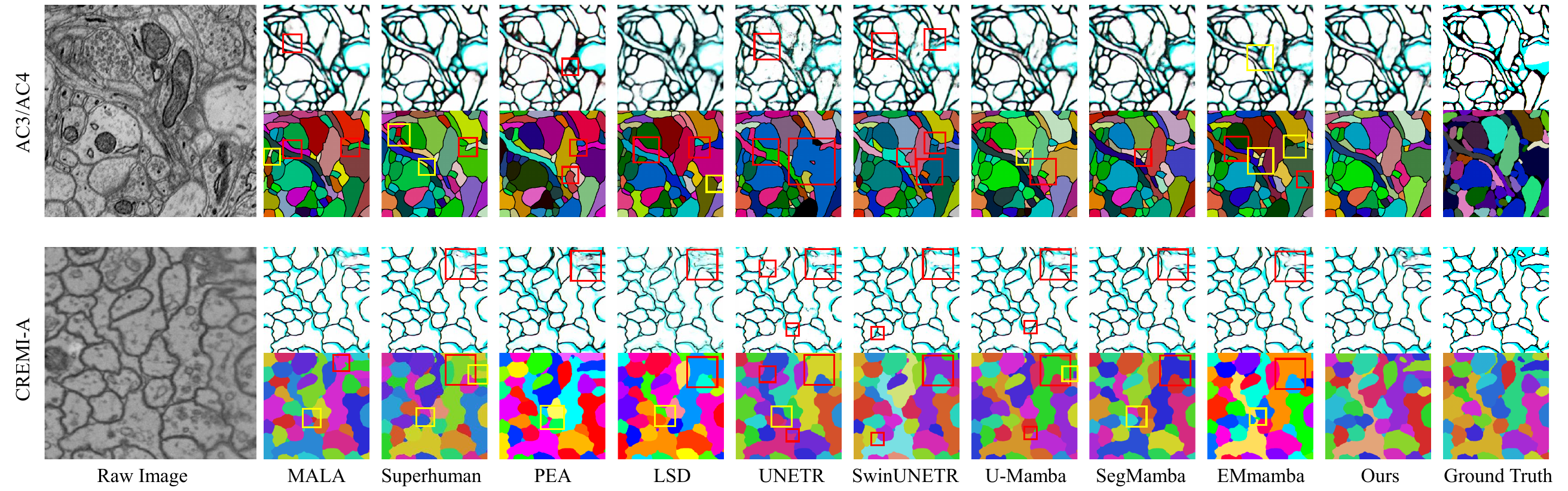} 
	\caption{2D visualization results on AC3/AC4 and CREMI-A data. The left 
		side shows the EM raw image, and the right side shows the affinity and 
		segmentation results. The red and yellow boxes indicate merge and split 
		errors, respectively}
	\label{fig4}
\end{figure*}

\noindent\textbf{AC3/AC4.}
This dataset \cite{AC2015} consists of two volumes cropped from the mouse 
somatosensory cortex imaged using scanning electron microscopy (SEM). The sizes 
of the AC3 and AC4 volumes are $256 \times 1024 \times 1024$ and 
$100 \times 1024 \times 1024$, respectively, with a resolution of 
$6 \times 6 \times 29\,\mathrm{nm}^3$. We use the top 80 slices of AC4 for 
training and the top 100 slices of AC3 for testing.

\noindent\textbf{CREMI.}
The CREMI dataset \cite{cremi2018} originates from the brain of an adult 
Drosophila fly, imaged using serial section transmission EM (ssTEM) at a 
resolution of $4 \times 4 \times 40\,\mathrm{nm}^3$. The dataset includes three 
labeled subsets (CREMI-A/B/C), each consisting of 125 consecutive slices. We 
use the top 60 slices of each subset for training and the bottom 50 slices for 
testing.

\noindent\textbf{FIB25.}
Unlike the previous two anisotropic datasets, the FIB25 dataset \cite{a:4} 
images the Drosophila brain using focused ion beam SEM (FIB-SEM) at an 
isotropic resolution of $8\,\mathrm{nm}$. Two volumes, each of size 
$520 \times 520 \times 520$, are used for training and testing, respectively.

\noindent\textbf{Kasthuri.}
The Kasthuri dataset \cite{AC2015} has a size of $1850 \times 10747 \times 
12895$ 
with a voxel resolution of $6 \times 6 \times 29\,\mathrm{nm}^3$. In this work, 
we 
cropped a subset of size $300 \times 4096 \times 4096$ from the Kasthuri 
dataset, 
guided by the spatial distribution of annotated neurons. The bounding box of 
this 
subset spans from $(1050, 6500, 3200)$ to $(1350, 10596, 7296)$. We used AC4 
for 
training and evaluated large-volume neuron segmentation on this subset.

\subsection{Metrics}
We use two widely adopted metrics to evaluate neuron segmentation performance 
in EM images: variation of information (VI) and adapted Rand error (ARAND). 
Lower values of these metrics indicate better segmentation quality.

\subsection{Implementation Details}
Each model is trained on a single NVIDIA V100 GPU with the Adam optimizer. 
All models are trained for 200,000 epochs with a fixed learning rate of 0.0001 
and a batch size of 2.

\subsection{Comparison with State-of-the-art Methods}
We compared several state-of-the-art methods, including CNN-based methods 
(MALA, Superhuman, PEA, and LSD), Transformer-based methods (UNETR and 
SwinUNETR), and Mamba-based methods (U-Mamba, SegMamba, and EMMamba). All 
methods were trained and evaluated using block shapes consistent with those 
reported in the original papers, with slight adjustments along the $z$-axis to 
accommodate the limited number of slices in the EM datasets. Specifically, MALA 
uses $[53, 268, 268]$, Superhuman, PEA, and LSD use $[18, 160, 160]$, UNETR and 
SwinUNETR use $[64, 96, 96]$, and U-Mamba, SegMamba, and EMMamba use $[32, 
128, 128]$. The block shape used in our method is $[18, 160, 160]$. During 
inference, we adopt a tiling strategy with overlapping blocks, where the 
overlap 
along each dimension is set to half of its corresponding block size. After 
predicting the affinity map, two standard post-processing techniques were 
applied: Waterz \cite{mala} and Multicut \cite{multicut}.

\noindent\textbf{Quantitative Analysis.}
The quantitative results, presented across Tables~\ref{tab1}, \ref{tab2}, and 
\ref{tab22}, demonstrate that NeuroMamba consistently achieves 
state-of-the-art performance across all evaluated benchmarks. As highlighted in 
Tab.~\ref{tab1}, on the particularly 
challenging CREMI-A benchmark, NeuroMamba yields substantial 
relative ARAND improvements of \textbf{22.4\%} and \textbf{21.7\%} with Waterz 
and Multicut post-processing, respectively. This 
robustness and generalization are further substantiated in 
Tab.~\ref{tab2}, where our approach secures state-of-the-art results on 
both isotropic (FIB25) and anisotropic 
datasets, underscoring its adaptability to varying data resolutions. The 
superior performance on the large-scale Kasthuri dataset (Tab.~\ref{tab22}) 
further affirms the scalability and suitability of our 
method for large-volume, real-world reconstruction tasks.

\noindent\textbf{Qualitative Analysis.}
The visual comparisons in Figs.~\ref{fig4} and \ref{fig5} provide qualitative 
evidence of our method's superior performance. In both 2D and 3D 
visualizations, reconstructions generated by NeuroMamba exhibit significantly 
improved overall continuity and accurately segment fine structures that prove 
challenging for competing methods. This significant improvement is directly 
attributable to the synergy between our core components: the BDFE module, which 
enhances feature representation in ambiguous boundary regions, and the SCFE 
module, which incorporates global contextual information for each neuron. This 
effective fusion of fine-grained boundary cues with global spatial connectivity 
is pivotal in achieving more complete and accurate neuron reconstructions.

\noindent\textbf{Model Complexity Comparison.}
As shown in Tab.~\ref{tab4}, we compare the model complexity of various 
state-of-the-art methods. The inference latency is measured as the time 
required to process a standardized volume of size $[64, 96, 96]$. Our model 
achieves superior performance while maintaining a parameter count and 
computational cost comparable to lightweight CNNs. Compared with other 
Transformer-based and Mamba-based methods, NeuroMamba requires fewer parameters 
and floating-point operations. The inference latency of NeuroMamba is higher 
than that of CNN-based methods but lower than that of most Transformer-based 
and Mamba-based methods.

\subsection{Ablation Studies and Analysis}
\begin{figure*}[htbp]
	\centering
	\includegraphics[width=1.0\textwidth]{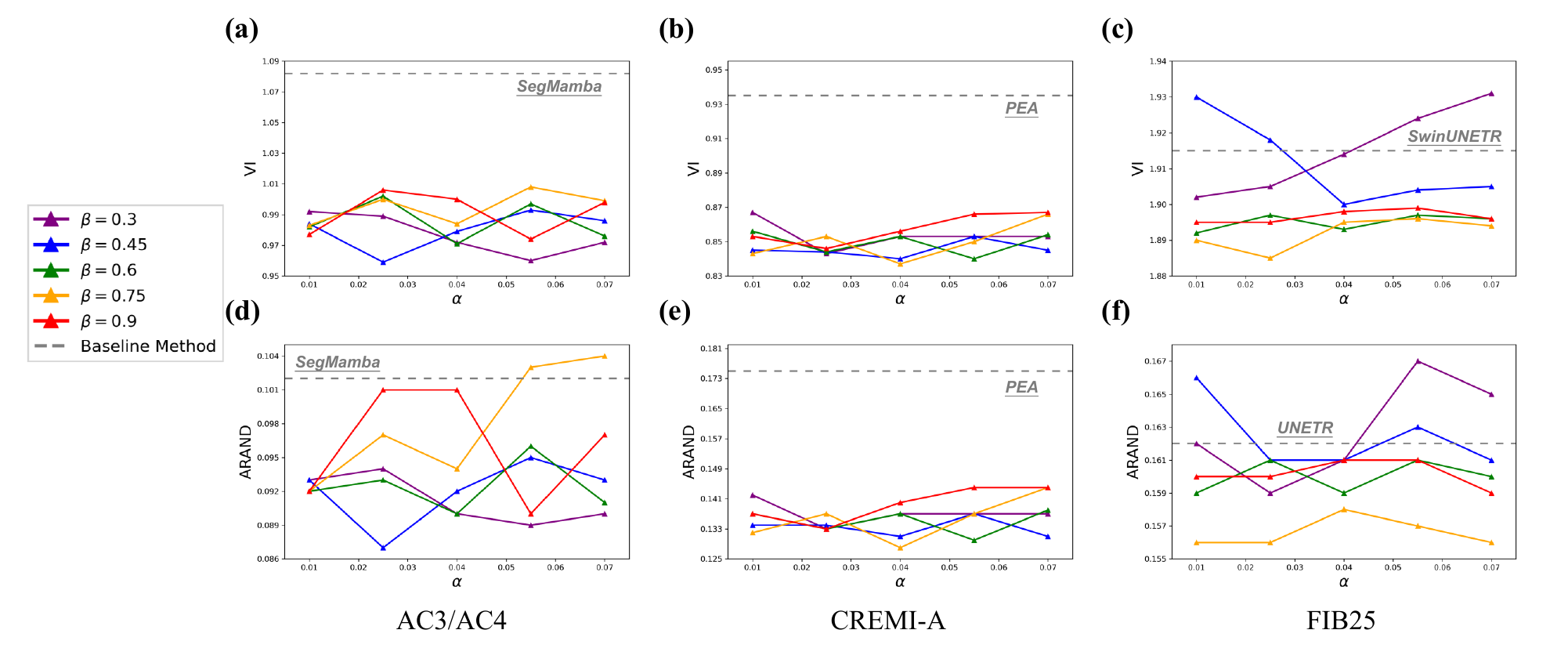} 
	\caption{We evaluate the performance of our model under different 
		hyperparameter configurations on three datasets. The top row reports 
		the VI 
		metric, while the bottom row presents the ARAND metric. The 
		best-performing 
		baseline method and its performance are highlighted in the figure.
	}
	\label{fig6}
\end{figure*}
To validate the effectiveness of our method, we conducted a series of ablation 
studies on different datasets.

\noindent\textbf{Effectiveness of Main Components.}
We first validated the 
effectiveness of the BDFE and SCFE modules. Tab.~\ref{tab3_2} presents the 
experimental 
results under different module configurations, demonstrating that an optimal 
neuron segmentation model requires the joint modeling of both global spatial 
continuous and local boundary discriminative information. Next, we examined the 
efficacy of the CFI module's information interaction. Specifically, we compared 
it against three commonly used information fusion methods: addition, 
element-wise multiplication, and concatenation. The results in 
Tab.~\ref{tab3_3} 
indicate 
that the information interaction method based on cross-modulation can 
dynamically adjust the weights of global spatial continuous and local boundary 
discriminative features, enabling the model to more flexibly handle neuron 
instances of varying scales.

\noindent\textbf{Effectiveness of Resolution-aware Modules.}
To validate that introducing resolution prior information enables the model to 
better adapt to data with different resolutions, we conducted ablation 
experiments on multiple datasets. As shown in Tab.~\ref{tab3_4}, our model 
achieves improved performance across different datasets after incorporating 
resolution prior information.

\noindent\textbf{Effectiveness of Mamba.}
Transformer-based methods require partitioning 3D blocks into patches, which 
neglects the spatial relationships between voxels within each block and leads 
to the loss of affinity information. Our proposed NeuroMamba models the 
relationships among all voxels without patch partitioning, effectively 
mitigating this issue. As shown in Tab.~\ref{tab3_5}, we first replace the 
transformer block in UNETR with the Mamba block to obtain UNEM. The comparative 
experiments demonstrate that the Mamba block can better capture long-range 
neuronal dependencies. However, its performance remains significantly inferior 
to that of NeuroMamba, which simultaneously models both global voxel-level 
information and local spatial proximity.

\noindent\textbf{Effectiveness of Strip Pooling.}
\begin{figure}[t]
	\centering
	\includegraphics[width=0.87\columnwidth]{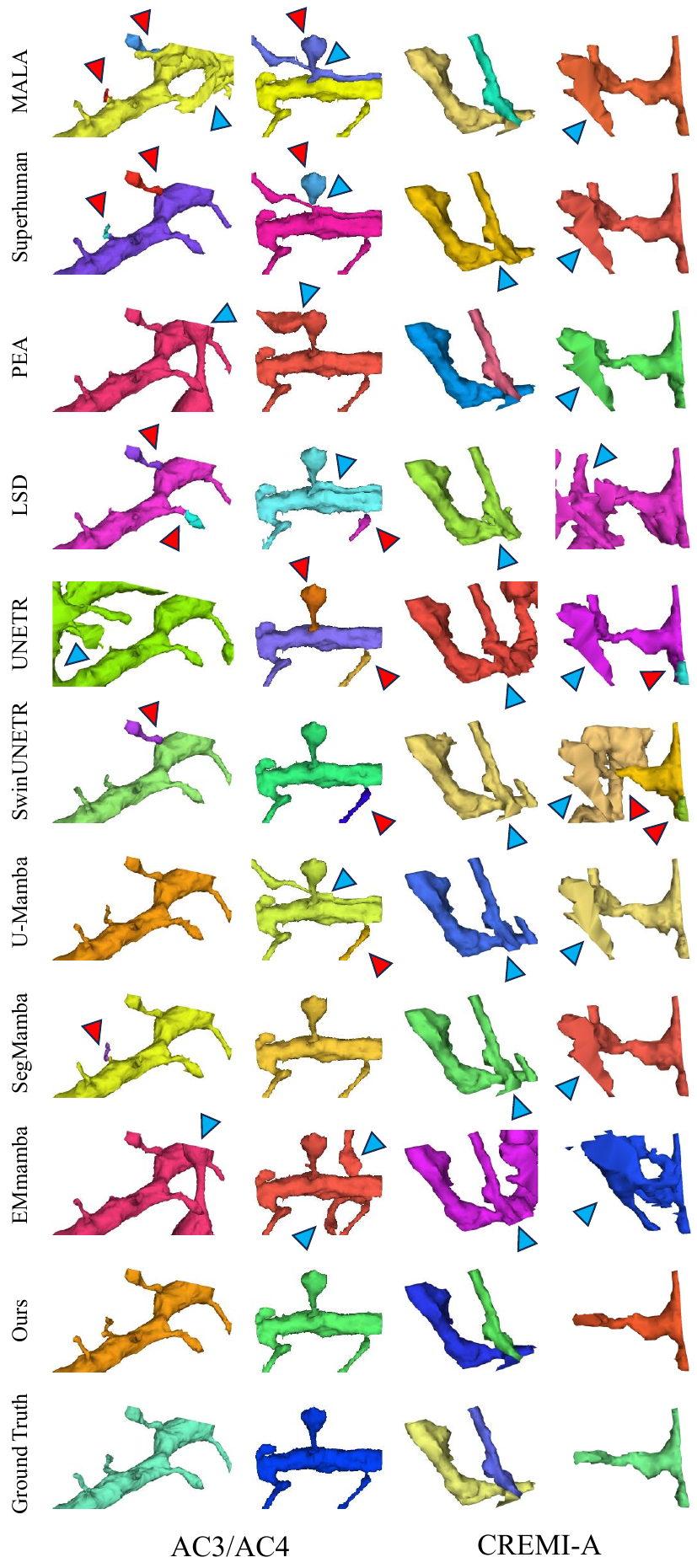} 
	\caption{3D visualization results on AC3/AC4 and CREMI-A data. The blue and 
		red arrows indicate merge and split errors, respectively.}
	\label{fig5}
\end{figure}
As shown in Tab.~\ref{tab3_6}, replacing strip pooling with square pooling in 
NeuroMamba leads to a decrease in performance. This finding indicates that 
strip pooling is better suited to the elongated morphological characteristics 
of neurons.

\noindent\textbf{Robustness Across Different Block Shapes.}
As shown in Tab.~\ref{tab5}, NeuroMamba achieves comparable performance on 
CREMI-A across different block shapes. In each block-shape configuration, 
NeuroMamba consistently outperforms its corresponding baseline.

\noindent\textbf{Different Scanning Mechanisms.} 
To demonstrate the suitability of our proposed cross-scan mechanisms for 
modeling transverse and axial features in 3D EM data, we conducted comparative 
experiments with unidirectional and bidirectional scanning strategies 
\cite{visionmamba, videomamba}. We further compared methods employing 
Dual-scale 
\cite{voxelmamba} and Multi-scale \cite{multimamba} scanning strategies, as 
well 
as the Hilbert scanning technique \cite{voxelmamba,mambaAD}, which has 
demonstrated strong performance in detection tasks. As shown in 
Tab.~\ref{tab6}, our proposed transverse-first and axial-first cross-scan 
mechanisms achieve the best performance. This may be attributed to the fact 
that the cross-scan strategy effectively mitigates the asymmetry introduced by 
one-dimensional unwrapping, while scanning in both transverse and axial 
directions allows the model to better capture structural information along 
multiple orientations.

\noindent\textbf{Ablation Experiments on Hyperparameter Tuning.} 
The SCFE module dynamically adjusts the weights $\lambda_1$ and $\lambda_2$ 
based on the resolution prior of the EM data, as shown in \ref{eq:9} and 
\ref{eq:10}. As shown in Fig.\ref{fig6}, we conducted experiments under 
multiple 
parameter settings across the three datasets. The parameter $\alpha$ is set to 
$\{0.01, 0.025, 0.04, 0.055, 0.07\}$, and $\beta$ is set to $\{0.3, 0.45, 0.6, 
0.75, 0.9\}$. Our model consistently outperforms the best-performing baseline 
in most parameter configurations, demonstrating its robustness to 
hyperparameter variation. Finally, we adopt $\alpha=0.04$ and $\beta=0.6$ as 
they yield the most stable performance across the three datasets.

\begin{table}[t]
	\centering
	\small
	\setlength{\tabcolsep}{1pt}
	\begin{tabular}{c|c|c}
		\toprule
		\multicolumn{1}{c|}{block shape} 
		& Best baseline (VI/ARAND)
		& Ours (VI/ARAND) \\ 
		\midrule     	
		\multicolumn{1}{c|}{[18,160,160]}  
		& 0.935 / 0.175 & 0.853 / 0.137 \\ 
		
		\multicolumn{1}{c|}{[64,96,96]}  
		& 1.066 / 0.224 & 0.845 / 0.134 \\  
		
		\multicolumn{1}{c|}{[32,128,128]}  
		& 1.003 / 0.201 & 0.851 / 0.133 \\ 
		
		\bottomrule
	\end{tabular}
	\caption{Results under different block shapes.}
	\label{tab5}
\end{table}

\begin{table}[t]
	\centering
	\small
	\renewcommand{\arraystretch}{0.9}
	\setlength{\tabcolsep}{4.0pt}
	\begin{tabular}{cc|cc|cc}
		\toprule
		\multicolumn{2}{c|}{\multirow{2}{*}{Methods}} &  
		\multicolumn{2}{c|}{Waterz} &
		\multicolumn{2}{c}{Multicut} \\ 
		\cmidrule{3-6} 
		\multicolumn{2}{c|}{} &
		\multicolumn{2}{c|}{$\mathrm{VI}\downarrow$ \qquad 
			$\mathrm{ARAND}\downarrow$} &
		\multicolumn{2}{c}{$\mathrm{VI}\downarrow$ \qquad 
			$\mathrm{ARAND}\downarrow$} \\ 
		\midrule     
		\multicolumn{2}{c|}{$Uni^*$}
		& 0.899 & 0.170 & 0.878 & 0.151 \\ 
		\multicolumn{2}{c|}{$Uni^\dagger$}  
		& 0.903 & 0.161 & 0.858 & 0.144 \\
		\multicolumn{2}{c|}{$Uni^*+Uni^\dagger$} 
		& 0.899 & 0.167 & 0.862 & 0.141 \\
		\multicolumn{2}{c|}{$Bi^*$}  
		& 0.907 & 0.172 & 0.890 & 0.157 \\ 
		\multicolumn{2}{c|}{$Bi^\dagger$} 
		& 0.906 & 0.168 & 0.883 & 0.150 \\
		\multicolumn{2}{c|}{$Bi^*+Bi^\dagger$} 
		& 0.910 & 0.159 & 0.876 & 0.147 \\
		\multicolumn{2}{c|}{$Cro^*$}  
		& 0.884 & 0.169 & 0.855 & 0.139 \\ 
		\multicolumn{2}{c|}{$Cro^\dagger$} 
		& 0.885 & 0.155 & 0.868 & 0.143 \\
		\multicolumn{2}{c|}{$Cro^*+Cro^\dagger$} 
		& \textbf{0.865} & \textbf{0.142}
		& \textbf{0.853} & \textbf{0.137} \\
		\multicolumn{2}{c|}{Dual-scale} 
		& 0.891 & 0.166 & 0.856 & 0.140 \\
		\multicolumn{2}{c|}{Multi-scale} 
		& 0.924 & 0.168 & 0.882 & 0.152 \\
		\multicolumn{2}{c|}{Hilbert} 
		& 0.872 & 0.146 & 0.883 & 0.151 \\
		
		\bottomrule
	\end{tabular}
	\caption{Ablation of different scanning mechanisms on CREMI-A. 'Uni' 
		represents Unidirectional Scan, 'Bi' represents Bidirectional Scan, and 
		'Cro' represents Cross-Scan. '$^*$' indicates Transverse-First, while 
		'$^\dagger$' indicates Axial-First.}
	\label{tab6}
\end{table}

\section{Conclusion}
In this paper, we have presented NeuroMamba, a novel multi-perspective 
framework for neuron segmentation in volumetric EM data. By identifying the 
complementary weaknesses of existing methods, we developed a synergistic 
architecture that integrates a channel-gated module (BDFE) for boundary 
discriminative features and a resolution-aware Mamba-based module (SCFE) to 
efficiently model long-range spatial continuity without patching. These 
multi-perspective features are dynamically fused via a cross-modulation 
mechanism. Extensive experiments on four public benchmarks demonstrate the 
superiority and robustness of our method, culminating in a relative 
improvement of 22.4\% in the ARAND metric on the challenging CREMI-A 
dataset, establishing a new state-of-the-art performance.

\bibliographystyle{IEEEtran}
\bibliography{tmi}

\end{document}